\documentclass[10pt,twocolumn,letterpaper]{article}

\usepackage{caption}
\usepackage{cvpr}
\usepackage{times}
\usepackage{epsfig}
\usepackage{graphicx,lipsum,afterpage}
\usepackage{amsmath}
\usepackage{amssymb}
\usepackage{multirow}
\usepackage{siunitx}
\usepackage{marvosym}
\usepackage{lipsum}
\usepackage{hyperref}
\hypersetup{pagebackref=true,breaklinks=true,letterpaper=true,colorlinks,bookmarks=false}

\cvprfinalcopy 

\def\cvprPaperID{2804} 

\begin{document}

\title{PANDA: A Gigapixel-level Human-centric Video Dataset}

\author{
Xueyang Wang\textsuperscript{1}\textsuperscript{*},
Xiya Zhang\textsuperscript{1}\textsuperscript{*},
Yinheng Zhu\textsuperscript{1}\textsuperscript{*},
Yuchen Guo\textsuperscript{1}\textsuperscript{*},
Xiaoyun Yuan\textsuperscript{1},
Liuyu Xiang\textsuperscript{1},
\\
Zerun Wang\textsuperscript{1},
Guiguang Ding\textsuperscript{1},
David Brady\textsuperscript{2},
Qionghai Dai\textsuperscript{1},
Lu Fang\textsuperscript{1\Letter}
\\
\textsuperscript{1}Tsinghua University
\hspace*{3em}\textsuperscript{2}Duke University
}

\twocolumn[{%
\renewcommand\twocolumn[1][]{#1}%
\maketitle
\vspace{-12mm}
\begin{center}
 \centering
\includegraphics[width=1\textwidth]{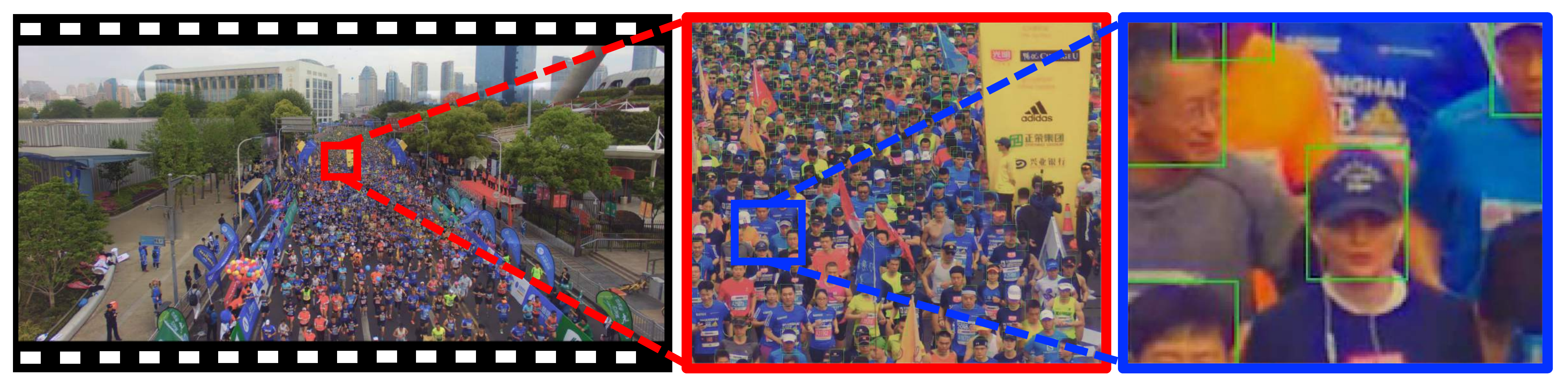}
 \captionof{figure}{A representative video \textit{Marathon} of PANDA dataset. The characteristic of joint wide field-of-view and high spatial resolution empowers the large-scale, long-term, and multi-object visual analysis.}
 \label{fig:fig_1_teaser}
\end{center}
}]



\begin{abstract}

\newcommand\blfootnote[1]{%
  \begingroup
  \renewcommand\thefootnote{}\footnote{#1}%
  \addtocounter{footnote}{-1}%
  \endgroup
}

\vspace{-3.5mm}
We present PANDA, the first gigaPixel-level humAN-centric viDeo dAtaset, for large-scale, long-term, and multi-object visual analysis. The videos in PANDA were captured by a gigapixel camera and cover real-world scenes with both wide field-of-view ($\sim${\SI{1}{\kilo\metre\squared}} area) and high-resolution details ($\sim$gigapixel-level/frame). The scenes may contain $4k$ head counts with over $100\times$ scale variation. PANDA provides enriched and hierarchical ground-truth annotations, including $15,974.6k$ bounding boxes, $111.8k$ fine-grained attribute labels, $12.7k$ trajectories, $2.2k$ groups and $2.9k$ interactions. We benchmark the human detection and tracking tasks. Due to the vast variance of pedestrian pose, scale, occlusion and trajectory, existing approaches are challenged by both accuracy and efficiency. Given the uniqueness of PANDA with both wide FoV and high resolution, a new task of interaction-aware group detection is introduced. We design a `global-to-local zoom-in' framework, where global trajectories and local interactions are simultaneously encoded, yielding promising results. We believe PANDA will contribute to the community of artificial intelligence and  praxeology by understanding human behaviors and interactions in large-scale real-world scenes. PANDA Website: http://www.panda-dataset.com. \blfootnote{\textsuperscript{* }These authors have contributed equally to this work.  \newline \hspace*{1.4em}\textsuperscript{\Letter} Corresponding author. Mail:  fanglu@sz.tsinghua.edu.cn. \newline \hspace*{1.4em}This work is supported in part by Natural Science Foundation of China (NSFC) under contract No. 61722209, 6181001011, 61971260 and U1936202, in part by Shenzhen Science and Technology Research and Development Funds (JCYJ20180507183706645).}

\end{abstract}

\section{Introduction}

It has been widely recognized that the recent conspicuous success of computer vision techniques, especially the deep learning based ones, rely heavily on large-scale and well-annotated datasets. For example, ImageNet~\cite{russakovsky2015imagenet} and CIFAR-10/100~\cite{TorralbaFF08} are important catalyst for deep convolutional neural networks~\cite{he2016deep, KrizhevskySH12}, Pascal VOC~\cite{everingham2010pascal} and MS COCO~\cite{lin2014microsoft} for common object detection and segmentation, LFW~\cite{huang2008labeled} for face recognition, and Caltech Pedestrians~\cite{dollar2011pedestrian} and MOT benchmark~\cite{milan2016mot16} for person detection and tracking. Among all these tasks, human-centric visual analysis is fundamentally critical yet challenging. It relates to many sub-tasks, e.g., pedestrian detection, tracking, action recognition, anomaly detection, attribute recognition etc., which attract considerable interests in the last decade~\cite{ren2015faster, cai2018cascade, lin2017focal, Wojke2017simple, long2018tracking, sun2019deep, cristani2011social, zanlungo2014pedestrian}. While significant progress has been made, \textbf{there is a lack of the long-term analysis of crowd activities at large-scale spatio-temporal range with clear local details}.

Analyzing the reasons behind, existing datasets~\cite{lin2014microsoft, dollar2011pedestrian, milan2016mot16, ferryman2009pets2009, ristani2016performance, blunsden2010behave} suffer an inherent trade-off between the wide FoV and high resolution. Taking the football match as an example, a wide-angle camera may cover the panoramic scene, yet each player faces significant scale variation, suffering very low spatial resolution. Whereas one may use a telephoto lens camera to capture the local details of the particular player, the scope of the contents will be highly restricted to a small FoV. Even though the multiple surveillance camera setup may deliver more information, the requisite of re-identification on scattered video clips highly affects the continuous analysis of the real-world crowd behaviour. All in all, existing human-centric datasets remain constrained by the limited spatial and temporal information provided. The problems of low spatial resolution \cite{lerner2007crowds, pellegrini2009you, ferryman2009pets2009}, lack of video information \cite{choi2014discovering,zen2010space,cristani2011social,bazzani2013social}, unnatural human appearance and actions \cite{alameda2015salsa, joo2015panoptic, ibrahim2016hierarchical}, and limited scope of activities with short-term annotations \cite{blunsden2010behave, list2005performance, choi2009they, oh2011large} lead to inevitable influence for understanding the complicated behaviors and interactions of crowd.


To address aforementioned problems, we propose a new gigaPixel-level humAN-centric viDeo dAtaset (PANDA). The videos in PANDA are captured by a gigacamera \cite{yuan2017multiscale,Brady2012MultiscaleGP}, which is capable of covering a large-scale area full of high resolution details. A representative video example of \textit{Marathon} is presented in Fig.~\ref{fig:fig_1_teaser}. Such rich information enables PANDA to be a competitive dataset with \textbf{multi-scale features: (1) globally wide field-of-view where visible area may beyond 1 km$^2$, (2) locally high resolution details with gigapixel-level spatial resolution, (3) temporally long-term crowd activities with $43.7k$ frames in total, (4) real-world scenes with abundant diversities in human attributes, behavioral patterns, scale, density, occlusion, and interaction.} Meanwhile, PANDA is provided with rich and hierarchical ground-truth annotations, with $15,974.6k$ bounding boxes, $111.8k$ fine-grained labels, $12.7k$ trajectories, $2.2k$ groups and $2.9k$ interactions in total.

Benefiting from the comprehensive and multiscale information, PANDA facilitates a variety of fundamental yet challenging tasks for image/video based human-centric analysis. We start with the most fundamental detection task. Yet detection on PANDA has to address both the accuracy and efficiency issues. The former one is challenged by the significant scale variation and complex occlusion, while the latter one is highly affected by the gigapixel resolution. Whereafter, the task of tracking is benchmarked. Equipped with the simultaneous large-scale, long-term and multi-object properties, our tracking task is heavily challenged, due to the complex occlusion as well as large-scale and long-term activity existing in real-world scenarios. Moreover, PANDA enables a distinct task of identifying the group relationship of the crowd in the video, termed as interaction-aware group detection. In this task, we propose a novel global-to-local zoom-in framework to reveal the mutual effects between global trajectories and local interactions. Note that these three tasks are inherently correlated. Although detection may bias to local high-resolution detail and tracking may focus on global trajectories, the former promotes the latter significantly. Meanwhile, the spatial-temporal trajectories deduced from detection and tracking serve for group analysis.

In summary, PANDA aims to contribute a standardized dataset to the community, for investigating new algorithms to understand the complicated crowd social behavior in large-scale real-world scenarios. The contributions are summarized as follows.
\begin{itemize}
    \item We propose a new video dataset with gigapixel-level resolution for human-centric visual analysis. It is the first video dataset endowed with wide FoV and high spatial resolution simultaneously, which is capable of providing sufficient spatial and temporal information from both global scene and local details. Complete and accurate annotations of location, trajectory, attribute, group and intra-group interaction information of crowd are provided.

    \item We benchmark several state-of-the-art algorithms on PANDA for the fundamental detection and tracking tasks. The results demonstrate that existing methods are heavily challenged from both accuracy and efficiency perspectives, 
    and indicate that it is quite difficult to accurately detect objects in a scene with significant scale variation and track objects that move continuously for a long distance under complex occlusion.
    
    \item We introduce a new visual analysis task, termed as interaction-aware group detection, based on the spatial and temporal multi-object interaction. A global-to-local zoom-in framework is proposed to utilize the multi-modal annotations in PANDA, including global trajectories, local face orientations and interactions. Promising results further validate the collaborative effectiveness of global scene and local details provided by PANDA.
\end{itemize}

By serving the visual tasks related to the long-term analysis of crowd activities at a large-scale spatial-temporal range, we believe PANDA will definitely contribute to the community for understanding the complicated behaviors and interactions of crowd in large-scale real-world scenes, and further boost the intelligence of unmanned systems.

\section{Related Work}


\subsection{Image-based Datasets}

The most representative human-centric task on image datasets is human (person or pedestrian) detection. The common object detection datasets, such as PASCAL VOC \cite{everingham2010pascal}, ImageNet \cite{russakovsky2015imagenet}, MS COCO \cite{lin2014microsoft}, Open Images \cite{kuznetsova2018open} and Objects365 \cite{Shao_2019_ICCV} datasets, are not initially designed for human-centric analysis, although they contain human object categories\footnote{Different terms are used in these datasets, such as ``person'', ``people'', and ``pedestrian''. We uniformly use ``human'' when there is no ambiguity.}. However, restricted by the narrow FoV, each image only contains limited number of objects, far from enough to describe the crowd behaviour and interaction.

\textbf{Pedestrian Detection.} Some pioneer representatives include INRIA \cite{dalal2005histograms}, ETH \cite{ess2008mobile}, TudBrussels \cite{wojek2009multi}, and Daimler \cite{enzweiler2008monocular}. Later,  Caltech \cite{dollar2011pedestrian}, KITTI-D \cite{geiger2012we}, CityPersons \cite{zhang2017citypersons} and EuroCity Persons \cite{braun2019eurocity} datasets with higher quality, larger scale, and more challenging contents are proposed. Most of them are collected via a vehicle-mounted camera through the regular traffic scenario, with limited diversity of pedestrian appearances and occlusions. The latest WiderPerson \cite{zhang2019widerperson} and CrowdHuman \cite{shao2018crowdhuman} datasets focus on crowd scenes with many pedestrians. Due to the trade-off between spatial resolution and field of view, existing datasets cannot provide sufficient high resolution local details if the scene becomes larger.


\textbf{Group Detection.} Starting with the free-standing conversational groups (FCGs) decades ago~\cite{erving1963behavior}, the subsequent works try to study the interacting persons characterized by mutual scene locations and poses, known as F-formations \cite{kendon1990conducting}. Representatives ones include IDIAP Poster \cite{hung2011detecting}, Cocktail Party \cite{zen2010space}, Coffee Break \cite{cristani2011social} and GDet \cite{bazzani2013social}. In \cite{choi2014discovering}, the problem of structure group along with a dataset is proposed, which defines the way people spatially interact with each other. 
Recently, pedestrian group Re-Identification (G-ReID) benchmarks like DukeMTMC Group \cite{lin2019group} and Road Group \cite{lin2019group} are proposed to match a group of persons across different camera views. However, these datasets only support position-aware group detection, lack of the important dynamic interactions. 

\subsection{Video-based Datasets}

\textbf{Pedestrian Tracking.} It locates pedestrians in a series of frames and find the trajectories of them. MOT Challenge benchmarks \cite{leal2015motchallenge, milan2016mot16} were launched to establish a standardized evaluation of multiple objects tracking algorithms. The latest MOT19 benchmark \cite{dendorfer2019cvpr19} consists of 8 new sequences with very crowded challenging scenes. Besides, some datasets are designed for specific applications, e.g., Campus \cite{robicquet2016learning} and VisDrone2018 \cite{zhu2018vision}, which are drone-platform-based benchmarks. PoseTrack \cite{andriluka2018posetrack} contains joint position annotations for multiple persons in videos. To increase the FoV for long-term tracking, a network of cameras is adopted, leading to the multi-target multi-camera (MTMC) tracking problem. MARS \cite{zheng2016mars}, DukeMTMC \cite{ristani2016performance} are representative ones. 

On the other hand, to investigate pedestrians in surveillance perspectives, UCY Crowds-by-Example \cite{lerner2007crowds}, ETH BIWI Walking Pedestrians \cite{pellegrini2009you}, Town Center \cite{benfold2011stable} and Train Station \cite{zhou2012understanding} are proposed for trajectory prediction, abnormal behaviour detection, and pedestrian motion analysis. PETS09 \cite{ferryman2009pets2009} was collected by eight cameras in a campus for person density estimation, people tracking, event recognition, etc. Recently, CUHK \cite{shao2014scene} and WorldExpo'10 \cite{zhang2016data} serve for evaluating the performance of crowd segmentation, crowd density, collectiveness, and cohesiveness estimation. However, these datasets are in insufficient of the richness and complexity of the scenes, and can hardly provide high resolution local details, which is critical to further analyze the human interactions in crowd.







\textbf{Interaction Analysis.} SALSA \cite{alameda2015salsa} contains uninterrupted multi-modal recordings of indoor social events with 18 participants for over 60 minutes. Panoptic Studio \cite{joo2015panoptic} uses 480 synchronized VGA cameras to capture social interactions, with 3D body poses annotated. 
BEHAVE \cite{blunsden2010behave}, CAVIAR \cite{list2005performance}, Collective Activity \cite{choi2009they} and Volleyball \cite{ibrahim2016hierarchical} are widely used datasets to evaluate human group activity recognition approaches. VIRAT \cite{oh2011large} is a real-world surveillance video dataset containing diverse examples of multiple types of complex visual events. However, for the sake of local details, the group interactions are usually restricted in small scenes or unnatural human behaviors.

\section{Data Collection and Annotation}

\begin{figure*}[hbtp]
\vspace{-3mm}
\centering
\includegraphics[width= \linewidth]{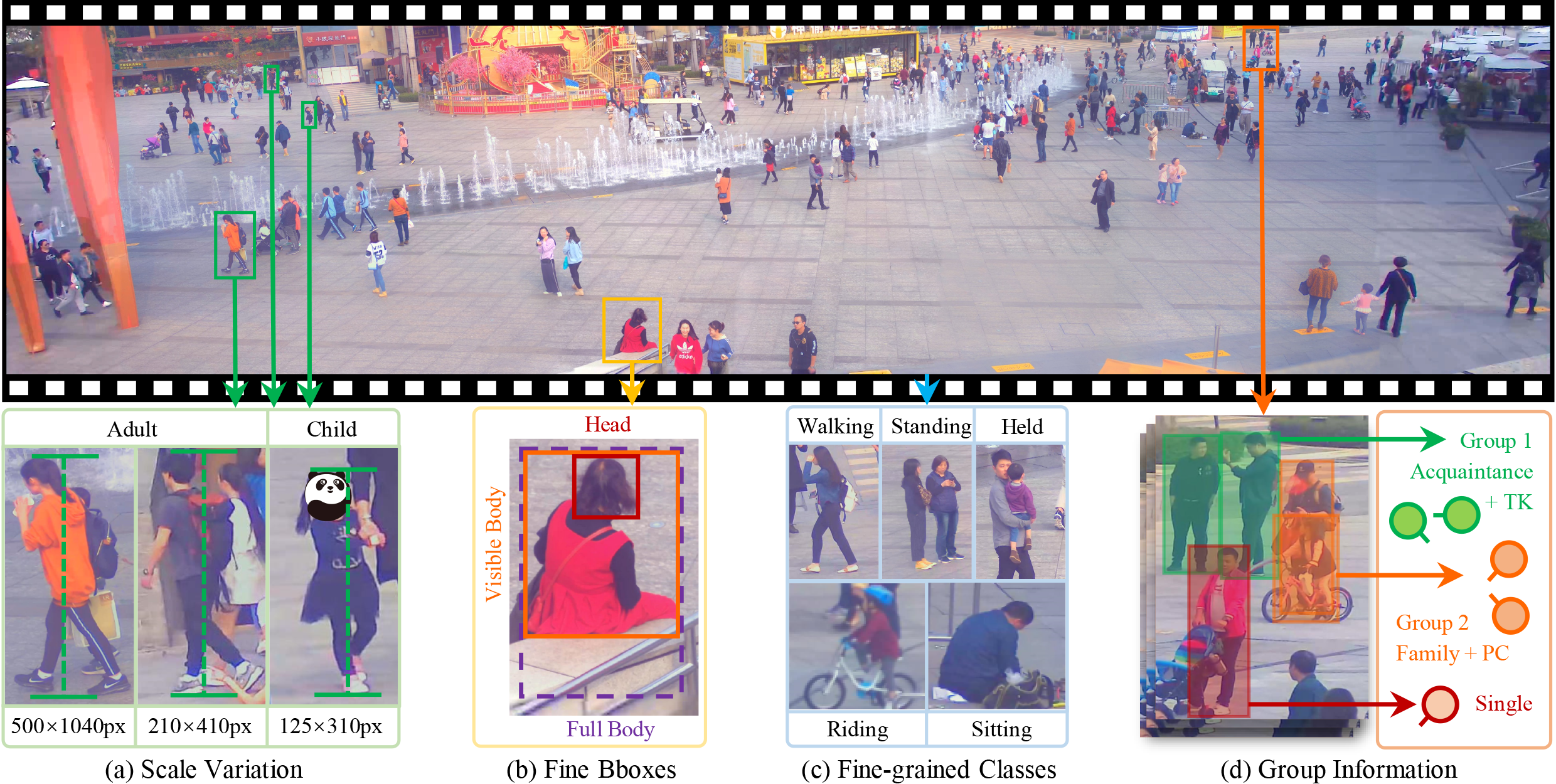}
\caption{Visualization of annotations in PANDA dataset. (a) The scale variation of pedestrians in a large-scale scene. (b) Three fine-grained bounding boxes on human body. (c) Five categories for human body postures. (d) Group information along with the intra-group interactions (TK=Talking, PC=Physical contact), where the circle and short line denote pedestrian and their face orientation.}
\vspace{-3mm}
\label{fig:demo}
\end{figure*}

\subsection{Data Collection and Pre-processing}


It is known that single camera based imaging suffers inevitable contradiction between wide FoV and high spatial resolution. The recently developed array-camera-based gigapixel videography techniques significantly boost the feasibility of high performance imaging \cite{Brady2012MultiscaleGP,yuan2017multiscale}. By designing the advanced computational algorithms, a number of micro-cameras work simultaneously to generate a seamless gigapixel-level video in realtime. As a result, the sacrifice in either field of view or spatial resolution can be eliminated. We adopt the latest gigacameras \cite{aquetiweb,yuan2017multiscale} to collect the data for PANDA, where the FoV is around 70 degree horizontally, and the video resolution reaches 25k$\times$14k, working in 30Hz. A representative video \textit{Marathon} in Fig. \ref{fig:fig_1_teaser} fully reflects the uniqueness of PANDA with both globally wide FoV and locally high resolution details. 

Currently, PANDA is composed by 21 real-world outdoor scenes\footnote{We are continuously collecting more videos to enrich our dataset. Note that all the data was collected in public areas where photography is officially approved, and it will be published under the Creative Commons Attribution-NonCommercial-ShareAlike 4.0 License \cite{license4.0}.}, by taking scenario diversity, pedestrians density, trajectory distribution, group activity, etc. into account. In each scene, we collected approximately 2 hours of 30Hz video as the raw data pool. Afterwards, around 3600 frames (approximately two minutes long segments) are extracted. For the images to be annotated, around 30 representative frames per video, 600 in total are selected, covering different crowd distributions and activities.

\subsection{Data Annotation}


Annotating PANDA images and videos faces the difficulty of full image annotation due to the gigapixel-level resolution. Herein, following the idea of divide-and-merge, the full image is partitioned into 4 to 16 subimages by considering the pedestrian density and size. After the labels are annotated on the subimages separately, the annotation results are mapped back to the full image. The objects cut by block borders are labeled with special status, which will be re-labeled after merging all blocks together. All labels are provided by a well trained professional annotation team.

\begin{table}
\begin{center}
\resizebox{\columnwidth}{!}{%
\begin{tabular}{|c|c|c|c|c|}
\hline
& Caltech  & CityPersons & PANDA  & PANDA-C     \\
\hline\hline
Res & 480P & 2048$\times$1024 & {\textbf{$\textgreater$25k$\times$14k}} & {\textbf{$\textgreater$25k$\times$14k}}  \\  
\#Im & \textbf{249.9k} & 5k & 555 & 45    \\ 
\#Ps & \textbf{289.4k} & 35.0k & 111.8k &  122.1k     \\ 
Den & 1.16 & 7.0 & \textbf{201.4} & \textbf{2,713.8}    \\
\hline
\end{tabular}
}
\end{center}
\caption{Pedestrian datasets comparison (statistics of CityPersons only contain public available training set). Res is the image resolution, \#Im is the total number of images, \#Ps is the total number of persons, Den denotes person density (average number of person per image) and PANDA-C is the PANDA-Crowd subset.}
\vspace{-5mm}
\label{tab:detection_comarison}
\end{table}

\subsubsection{Image Annotation}
PANDA has 600 well annotated images captured from 21 diverse scenes for the multi-object detection task. Among them, PANDA-Crowd subset are composed by 45 images labeled with human head bounding boxes, which are selected from 3 extremely crowded scenes that full of headcounts. The remaining 555 images from 18 real-world daily scenes own 111k pedestrians in total, labeled with head point, head bounding box, visible body bounding box, and the estimated full body bounding box close to the border of the pedestrian. For the crowd that are too far or too dense to be individually distinguished, the glass reflected persons, and the persons with more than 80\% occluded area are marked as `ignore' and disabled for benchmarking. 

Fig.~\ref{fig:demo} presents a typical large-scale real-world scene \textit{OCT Harbour} in PANDA, where the crowd shows a significant diversity in the scale, location, occlusion, activity, interaction, and so on. Beside the fine bounding boxes in (b), each pedestrian is further assigned a fine-grained label showing the detailed attributes in (c). Five categories are used, i.e., walking, standing, sitting, riding, and held in arms (for child), based on the daily postures. Pedestrians whose key parts occluded are marked as `unsure'. The `riding' label is further subdivided into bicycle rider, tricycle rider and motorcycle rider. Another detailed attribute is termed as `child' or `adult', distinguished from the appearance and behavior, as shown in (a). 

\begin{figure}[htbp]
\centering
\includegraphics[width=\linewidth]{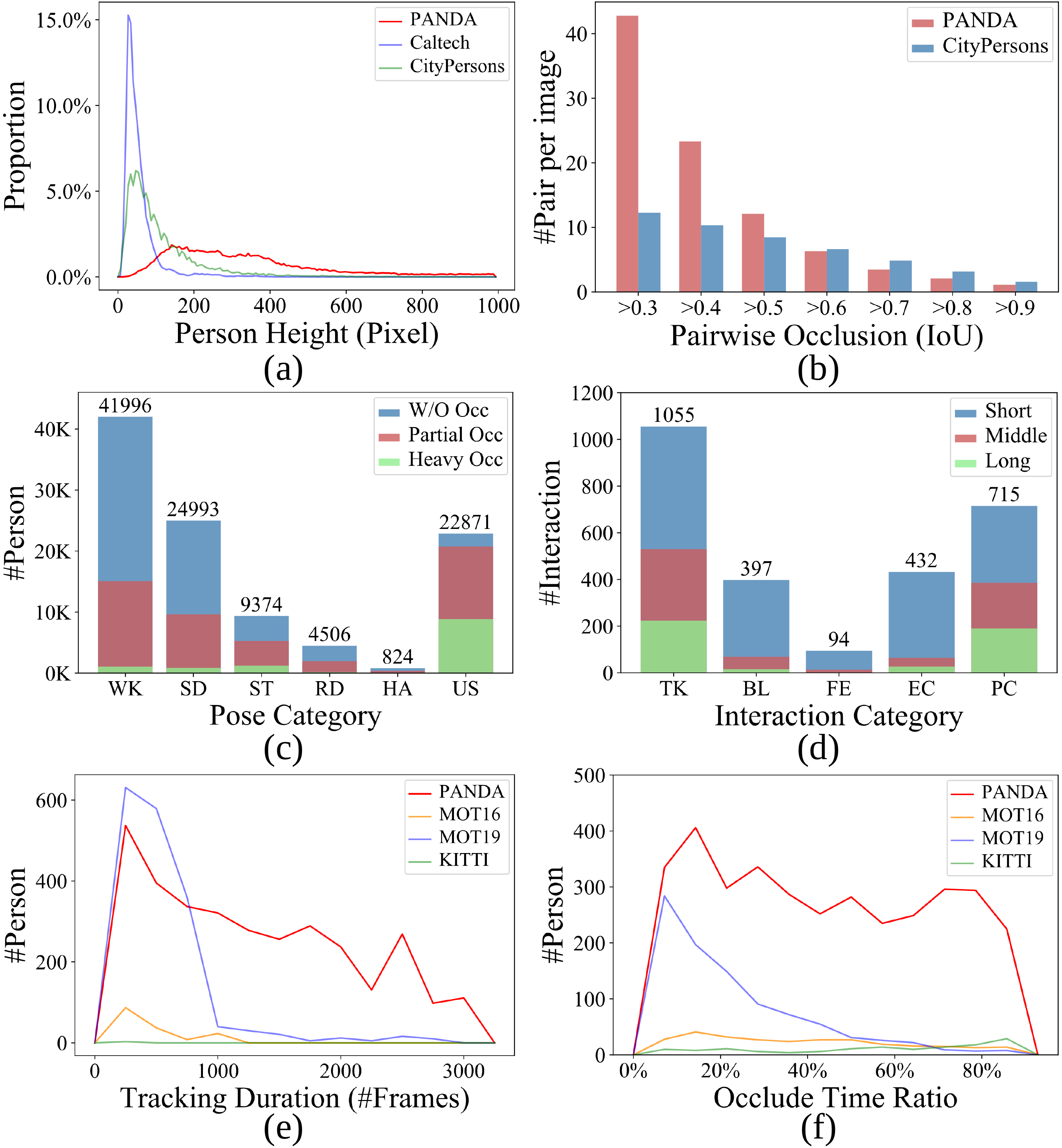}
\caption{(a) Distribution of person scale (height in pixel). (b) Distribution of the number of person pairs with different occlusion (measured by IoU) threshold per image. (c) Distribution of persons' pose labels in PANDA (WK=Walking, SD=Standing, ST=Sitting, RD=Ridding, HA=Held in arms, US=Unsure; The visible ratio is divided into W/O Occ (\textgreater 0.9), Partial Occ (0.5 - 0.9), and Heavy Occ (\textless 0.5)). (d) Distribution of categories and duration of inter-group interactions in PANDA (PC=Physical contact, BL=Body language, FE=Face expressions, EC=Eye contact, TK=Talking; The duration is divided into Short ($\textless$ 10s), Middle (10s - 30s), and Long ($\geq$ 30s)). (e) Distribution of person tracking duration. (f) Distribution of person occluded time ratio. The comparisons in (a), (b), (e) and (f) are limited to training sets.}
\vspace{-3mm}
\label{fig:sts1}
\end{figure}

The comparisons with the representative Caltech \cite{dollar2011pedestrian} and CityPersons image datasets \cite{zhang2017citypersons} are provided quantitatively (Tab.~\ref{tab:detection_comarison}) and statistically (Fig.~\ref{fig:sts1}). From Tab.~\ref{tab:detection_comarison}, each image of PANDA owns gigapixel-level resolution, which is around 100 times of existing datasets. 
Although the number of images is much smaller than other datasets, benefiting from the joint high resolution and wide FoV, PANDA has much higher pedestrian density per image than others especially in the extremely crowded PANDA-Crowd, and maintaining the total number of pedestrian in PANDA comparable to Caltech, 

Some detailed statistics about image annotation are shown in Fig.~\ref{fig:sts1}. In particular, Fig.~\ref{fig:sts1}(a) shows the distribution of person scale in pixel of PANDA, Caltech and CityPersons. As we can see, the height of persons in Caltech and CityPersons is mostly between 50px and 300px due to the limited spatial resolution, while PANDA has more balanced distribution from 100px to 600px. The larger scale variation in PANDA necessitates powerful multi-scale detection algorithms. In Fig.~\ref{fig:sts1}(b), the pairwise occlusion between persons measured by bounding box IoU of PANDA and CityPersons is given.
The fine-grained label statistics for different poses and occlusion conditions are summarized in Fig.~\ref{fig:sts1}(c). 

\subsubsection{Video Annotation}
Video annotation pays more attention on the labels revealing the activity/interaction. In addition to the bounding box of each person, we also label the face orientation (quantified into eight bins) and the occlusion ratio (without, partial and heavy). For pedestrians who are completely occluded for a short time, we label a virtual body bounding box and mark it as `disappearing'. MOT annotations are available for all the videos in PANDA except for PANDA-Crowd.


\begin{table}
\begin{center}
\resizebox{\columnwidth}{!}{%
\begin{tabular}{|c|c|c|c|c|}
\hline
& KITTI-T & MOT16 & MOT19 & PANDA \\
\hline\hline
Res & 1392$\times$512 & 1080p & 1080p & \textbf{\textgreater25k$\times$14k} \\
\#V & \textbf{20} & 14 & 8 & 15 \\
\#F & 19.1k & 11.2k & 13.4k & \textbf{43.7k} \\
\#T & 204 & 1.3k & 3.9k & \textbf{12.7k} \\
\#B & 13.4k & 292.7k & 2,259.2k & \textbf{15,480.5k} \\
Den & 0.7 & 26.1 & 168.6 & \textbf{354.6} \\
\hline
\end{tabular}
}
\end{center}
\caption{Comparison of multi-object tracking datasets (statistics of KITTI only contain public available training set). Res means video resolution. \#V, \#F, \#T and \#B denote the number of video clips, video frames, tracks and bounding boxes respectively. Den means Density (average number of person per frame).}
\vspace{-5mm}
\label{tab:mot_comparison}
\end{table}

The comparisons with KITTI-T \cite{geiger2012we} and MOT \cite{dendorfer2019cvpr19} video datasets are provided quantitatively (Tab.~\ref{tab:mot_comparison}) and statistically (Fig.~\ref{fig:sts1}). Apparently, PANDA is competitive with the largest number of frames, tracks and bounding boxes\footnote{Since the moving speed of pedestrians is relatively slow and stable, and the posture of pedestrians rarely changes rapidly and dramatically, we label them sparsely on every $k$ frames ($k = 6$ to $15$ based on the scene content) from the perspective of labeling cost. Here we compare the number of bounding boxes after linear interpolation to the original frame rate.}. Moreover, in Fig.~\ref{fig:sts1}(e), we show the distribution of tracking duration of different datasets. It demonstrates that the tracking duration in PANDA is many times longer than than KITTI-T and MOT because PANDA has wider FoV. This property makes PANDA an excellent dataset for large-scale and long-term tracking. Moreover, we also investigate the duration that each person is occluded and summarize the distribution in Fig.~\ref{fig:sts1}(f). It shows that more tracks in PANDA suffer from partial or heavy occlusions, in both absolute number and relative portion, making the tracking task more challenging. 

For group annotation, the advance of PANDA with wide-FoV global information, high-resolution local details and temporal activities ensures more reliable annotations for group detection. Unlike existing group-based datasets that focus on either the similarity of global trajectories \cite{pellegrini2009you} or the stability of local spatial structure \cite{choi2014discovering}, we utilize the social signal processing \cite{Vinciarelli2009SocialSP} to label the group attributes at the interaction level. 

More specifically, with the annotated bounding boxes, we firstly label the group members based on scene characteristics and social signals such as interpersonal distance \cite{duke1972new} and interaction \cite{Vinciarelli2009SocialSP}. Afterwards, each group is assigned an category label denoting the relationship, such as acquaintance, family, business, etc., as shown in Fig.~\ref{fig:demo}(d). To enrich the features for group identification, we further label the interactions between members within the group, including the interaction category (including physical contact, body language, face expressions, eye contact and talking; multi-label annotation) and its begin/end time. The distribution and duration of interactions are shown in Fig.~\ref{fig:sts1}(d). The mean duration of interaction is 518 frames (17.3s). To avoid overly subjective or ambiguous cases, three rounds of cross-checking are performed.

\section{Algorithm Analysis}


We consider three human-centric visual analysis tasks on PANDA. The first is pedestrian detection, which biases local visual information. The second is multi-pedestrian tracking. In this task, global visual clues from different regions are taken into consideration. Based on these two well-defined tasks, we introduce the interaction-aware group detection task. In this task, both global trajectories and local interactions between persons are necessary. 

\subsection{Detection}

\begin{table}[t]
\begin{center}
\resizebox{\columnwidth}{!}{%
\begin{tabular}{|c|c|c|c|c|c|c|c|}
\hline
\multirow{2}{*}{ } & \multirow{2}{*}{Sub} & \multicolumn{2}{c|}{Visible Body} & \multicolumn{2}{c|}{Full Body} & \multicolumn{2}{c|}{Head} \\ \cline{3-8} 
 & & AP$_{.50}$ & AR & AP$_{.50}$ & AR & AP$_{.50}$ & AR \\
\hline\hline
\multirow{3}{*}{FR \cite{ren2015faster}}  & S & 0.201 & 0.137 & 0.190 & 0.128 & 0.031 & 0.023 \\
 & M & 0.560 & 0.381 & 0.552 & 0.376 & 0.157 & 0.088 \\
 & L & 0.755 & 0.523 & 0.744  & 0.512 & 0.202 & 0.105 \\ \hline
\multirow{3}{*}{CR \cite{cai2018cascade}} & S & 0.204 & 0.140 & 0.227 & 0.160 & 0.028 & 0.018 \\
 & M & 0.561 & 0.388 & 0.579 & 0.384 & 0.168 & 0.091 \\
 & L & 0.747 & 0.532 & 0.765 & 0.518 & 0.241 & 0.116 \\ \hline
\multirow{3}{*}{RN \cite{lin2017focal}} & S & 0.171 & 0.121 & 0.221 & 0.150 & 0.023 & 0.018 \\
 & M & 0.547 & 0.370 & 0.561 & 0.360 & 0.143 & 0.081 \\
 & L & 0.725 & 0.482 & 0.740 & 0.479 & 0.259 & 0.149 \\
\hline
\end{tabular}
}
\end{center}
\caption{Performance of detection methods on PANDA. FR, CR, and RN denote Faster R-CNN, Cascade R-CNN and RetinaNet respectively. Sub means subset of different target sizes, where Small, Middle, and Large indicate object size being $< 32\times32$, $32\times32-96\times96$, and $>96\times96$.}
\vspace{-5mm}
\label{tab:det_result}
\end{table}

Pedestrian detection is a fundamental task for human-centric visual analysis. The extremely high resolution of PANDA makes it possible to detect pedestrians from a long distance. However, the significant variance in scale, posture, and occlusion severely degrade the detection performance. In this paper, we benchmarked several state-of-the-art detection algorithms on PANDA\footnote{For 18 ordinary scenes, 13 scenes are used for training and 5 scenes for testing. For 3 extremely crowded scenes, 2 scenes are for training and 1 scene for testing.}.

\textbf{Evaluation metrics.} For evaluation, we choose $AP_{.50}$ and $AR$ as metrics: $AP_{.50}$ is the average precision at $IoU=0.50$ and $AR$ is the average recall with $IoU$ ranging in $[0.5, 0.95]$ with a stride of 0.05. 

\textbf{Baseline detectors.} We choose Faster R-CNN \cite{ren2015faster}, Cascade R-CNN \cite{cai2018cascade} and RetinaNet \cite{lin2017focal} as our baseline detectors with ResNet101 \cite{he2016deep} backbone. The implementation is based on  \cite{mmdetection}. To train the gigapixel images on our network, we resize the original size image into multiple scales and partition the image into blocks with appropriate size as neural network input. For the objects cut by block borders, we retain them if the preserved area overs $50\%$.
Similarly, for evaluation, we resize the original image into multiple scales and use the sliding window approach to generate proper size blocks for the detector.
For a better analysis of detector performance and limitations, we split test results into subsets according to the object size.





\begin{figure}[t]
\centering
\includegraphics[width= \linewidth]{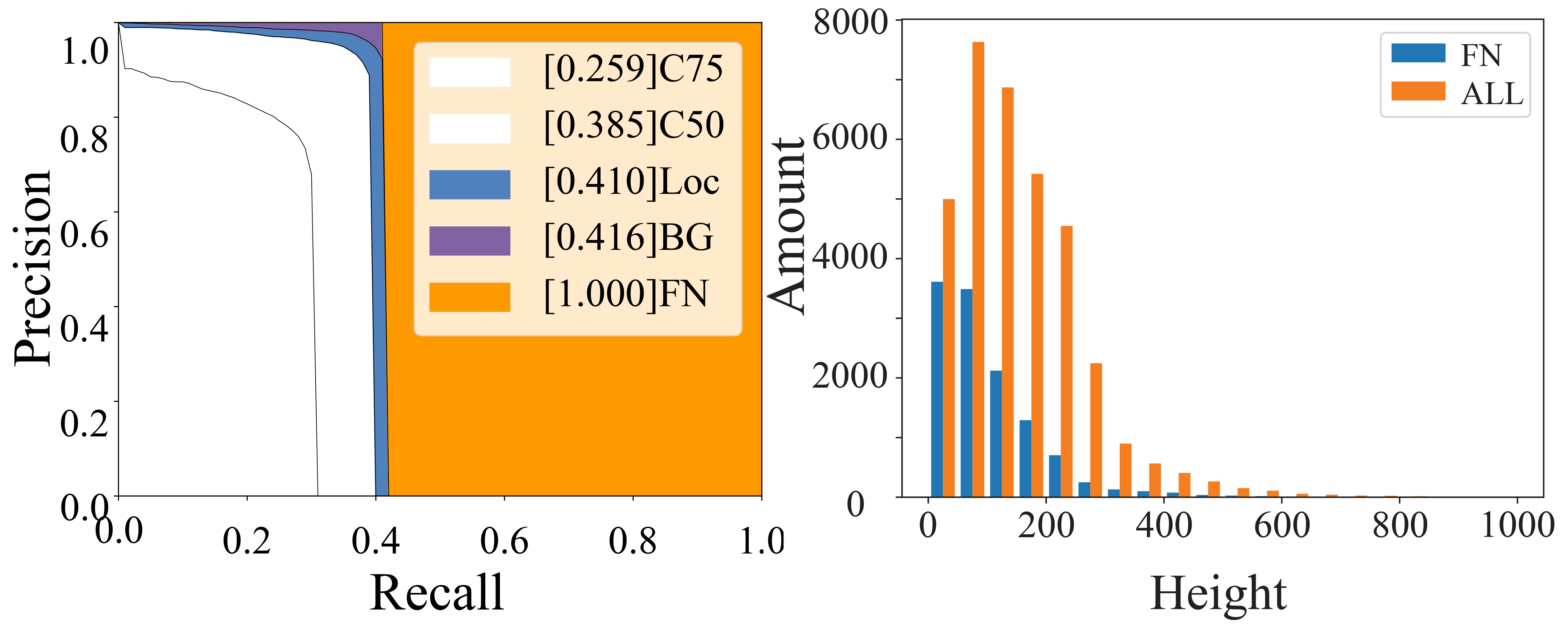}
\caption{Left: False analysis for Faster R-CNN on Visible Body. C75, C50, Loc and BG denote PR-curve at IoU=0.75, IoU=0.5, ignoring localization errors and ignoring background false positives, respectively. Right: False negative instances (FN) v.s. All instances (ALL) in terms of person height (in pixel) distribution for Faster R-CNN on Visible Body.}
\vspace{-5mm}
\label{fig:det_false}
\end{figure}

\textbf{Results.}
We train these 3 detectors from the COCO pre-trained weights and evaluate them on three tasks: visible body, full body, and head detection.
As shown in Tab.~\ref{tab:det_result}, Faster R-CNN, Cascade R-CNN and RetinaNet show the difficulty in detecting small objects, resulting in very low precision and recall. We also apply false analysis on visible body using Faster R-CNN, as illustrated in Fig.~\ref{fig:det_false} left. We can observe that the huge amount of false negatives is the most severe factor limiting the performance of the detector. 
We further analyze the height distribution of the false negative instances in Fig.~\ref{fig:det_false} right. The results indicate false negative caused by missing detection of small objects is the main reason for poor recall. 
According to the results, it seems quite difficult to accurately detect objects in a scene with very large scale variation (most 100$\times$ in PANDA) by the single detector based on existing architectures. More advanced optimization strategies and algorithms are highly demanded for the detection task on extra-large images with large object scale variation, such as scale self-adaptive detectors and efficient global-to-local multi-stage detectors.

\begin{figure*}[htbp]
\centering
\includegraphics[width=\linewidth]{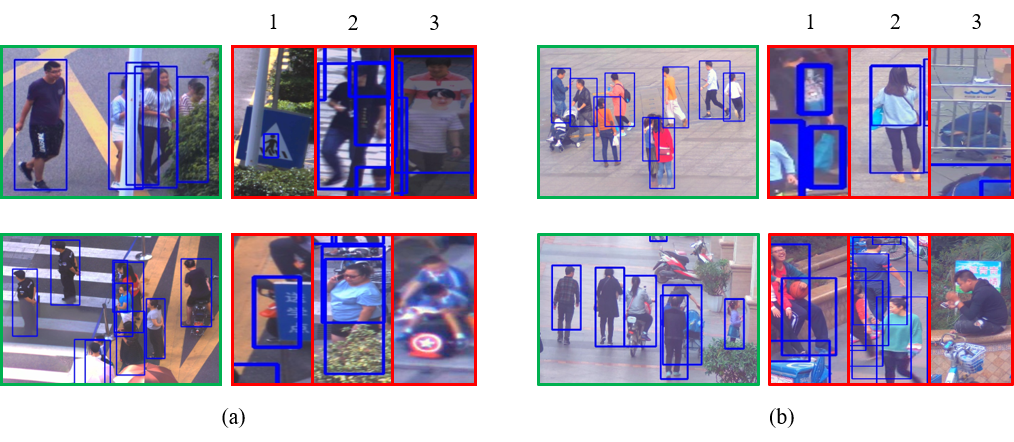}
\caption{Success detection cases (green) and failure detection cases (red). (a) Cascade R-CNN on Full Body. (b) Faster R-CNN on Visible Body. The failure cases can be summarized into three types: (1) confusion detection of the human-like objects; (2) duplicated detection on a single instance induced by the sliding window strategy; (3) missing detection of the human body with irregular size and scale due to occlusion or curled pose.}
\vspace{-5mm}
\label{fig:det_case}
\end{figure*}

Fig.~\ref{fig:det_case} depicts the representative failure and success cases of our detection results. As shown in the success cases, our detectors are capable to detect human body with various scale and poses by utilizing the local high-resolution visual feature. On the other hand, there are three types of failure cases: 1) confusion detection of the human-like objects; 2) duplicated detection on a single instance induced by the sliding window strategy; 3) missing detection of the human body with irregular size and scale due to occlusion or curled pose. These representative failure cases demonstrate the data diversity of our dataset that still has large room for improvement of the detection algorithms.

\subsection{Tracking}

Pedestrian tracking aims to associate pedestrians at different spatial positions and temporal frames. The superior properties of PANDA make it naturally suitable for long-term tracking. Yet the complex scenes with crowded pedestrian impose various challenges as well. 

\begin{table}
\begin{center}
\resizebox{\columnwidth}{!}{%
\begin{tabular}{|c|c|c|c|c|c|c|c|}
\hline
T & D & MOTA$\uparrow$ & MOTP$\uparrow$ & IDF1$\uparrow$ & FAR$\downarrow$ & MT$\uparrow$\\
\hline\hline
\multirow{3}{*}{DS \cite{Wojke2017simple}} & FR & \textbf{25.53} & 76.67 & 21.14 & 20.45 & 762\\ 
 & CR & 24.35 & 76.31 & 21.39 & 15.59 & 661 \\ 
 & RN & 16.36 & 78.0 & 15.16 & 4.32 & 259 \\ \hline
\multirow{3}{*}{DAN \cite{sun2019deep}} & FR & 25.06 & 74.81 & \textbf{21.85} & 25.95 & \textbf{826}\\
 & CR & 24.24 & 78.55 & 20.13 & 12.42 & 602 \\
 & RN & 15.57 & 79.90 & 13.43 & 3.33 & 227 \\ \hline
\multirow{3}{*}{MD \cite{long2018tracking}} & FR & 13.51 & 78.82 & 14.92 & 6.52 & 257\\ 
 & CR & 13.54 & 80.25 & 14.89 & 4.41 & 255 \\
 & RN & 10.77 & \textbf{80.62} & 11.86 & \textbf{1.90} & 162 \\
\hline
\end{tabular}
}
\end{center}
\caption{Performance of multiple object tracking methods on PANDA. T is tracker, D is detector, DS, DAN and MD denote the DeepSORT \cite{Wojke2017simple}, DAN \cite{sun2019deep} and MOTDT \cite{long2018tracking} trackers, respectively. $\uparrow$ denotes higher is better and vice versa.}
\vspace{-5mm}
\label{tab:mot_results}
\end{table}

\textbf{Evaluation metrics.} To evaluate the performance of multiple person tracking algorithms, we adopt the metrics of MOTChallenge \cite{leal2015motchallenge, milan2016mot16}, including MOTA, MOTP, IDF1, FAR, MT and Hz. Multiple Object Tracking Accuracy (MOTA) computes the accuracy considering three error sources: false positives, false negatives/missed targets and identity switches. Multiple Object Tracking Precision (MOTP) takes into account the misalignment between the groundtruth and the predicted bounding boxes. ID F1 score (IDF1) measures the ratio of correctly identified detections over the average number of ground-truth and computed detections. False alarm rate (FAR) measures the average number of false alarms per frame. The mostly tracked targets (MT) measures the ratio of ground-truth trajectories that are covered by a track hypothesis for at least 80\% of their respective life span. The Hz indicates the processing speed of the algorithm. For all evaluation metrics, except FAR, higher is better. 

\textbf{Baseline trackers.} Three representative algorithms DeepSORT \cite{Wojke2017simple}, DAN \cite{sun2019deep} and MOTDT \cite{long2018tracking} are evaluated. All of them follow the tracking-by-detection strategy. In our experiments, the bounding boxes are generated from 3 detection algorithms \cite{ren2015faster, cai2018cascade, lin2017focal} in the previous subsection. For the sake of fairness, We use the same pre-trained weights on the COCO dataset and detection threshold scores (0.7) for them. Default model parameters provided by the authors are used for evaluating three trackers.


\textbf{Results.} Tab.~\ref{tab:mot_results} shows the results of DeepSORT \cite{Wojke2017simple}, MOTDT \cite{long2018tracking} and DAN \cite{sun2019deep} on PANDA. 
The time cost to process a single frame is 18.36s (0.054Hz), 19.13s (0.052Hz), 8.29s (0.121Hz) for DeepSORT, MOTDT and DAN, respectively. DeepSORT and DAN show similar performance, but DAN is more efficient. MOTDT shows better bounding box alignment according to MOTP and FAR. DAN leads on IDF1 and MT, implying its stronger capability to establish correspondence between the detected objects in different frames.
The experimental results also demonstrate the challenge of PANDA dataset. The best MOTA for DeepSORT, DAN and MOTDT on MOT16 are 61.4, 52.42 and 47.6, while drop more than half on PANDA (The maximum MOTA is only 25.53).
With regards to object detectors, Faster R-CNN performs the best and Cascade R-CNN shows similar performance. Whereas the performance of RetinaNet is relatively poor except MOTP and FAR, the reason is that RetinaNet has low recall under confidence threshold 0.7 for detection results.

\begin{figure}[t]
\centering
\footnotesize
\includegraphics[width= \linewidth]{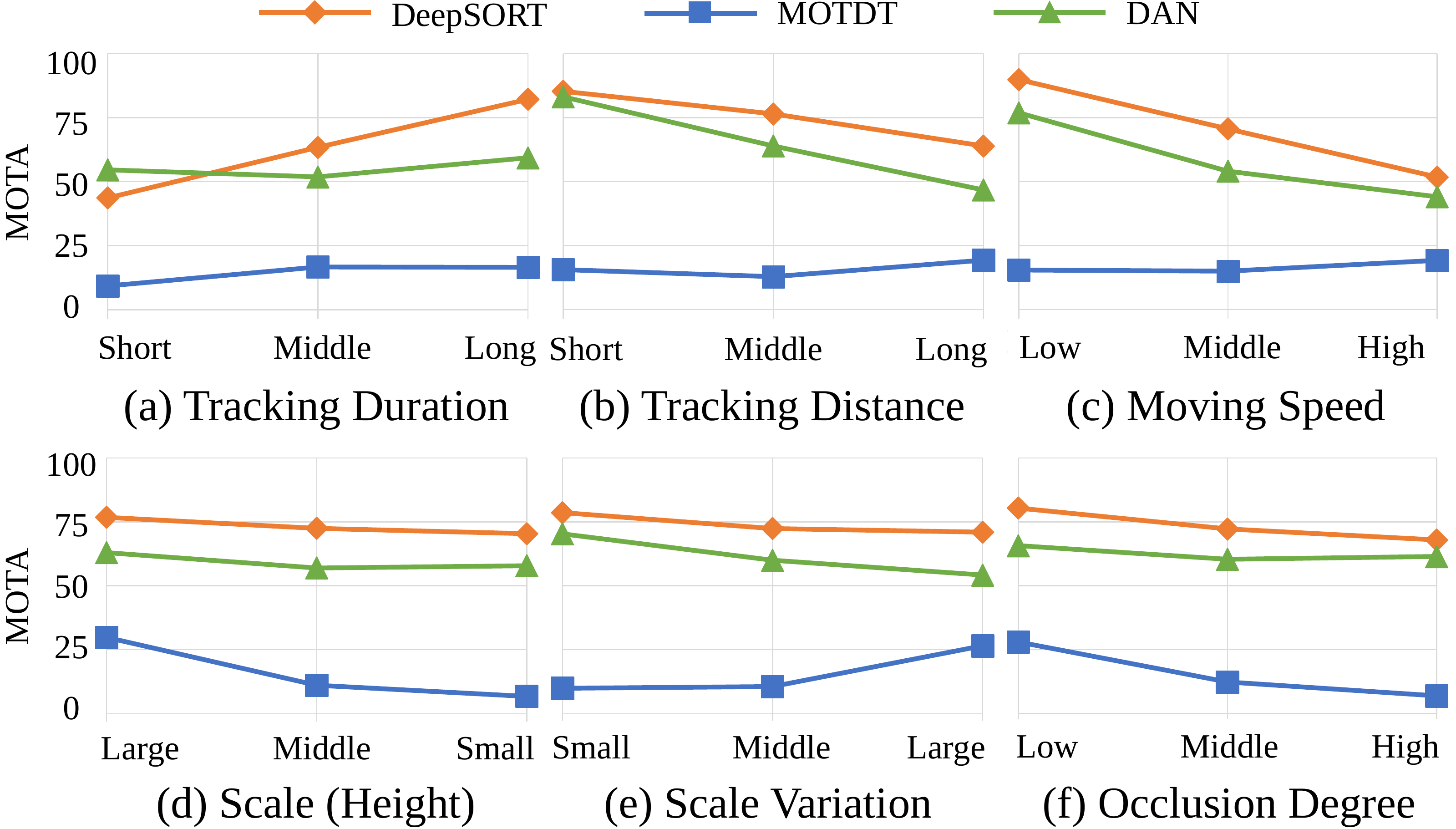}
\caption{Influence of target properties on tracker's MOTA. We divided the pedestrian targets into 3 subsets from easy to hard for each property.}
\vspace{-5mm}
\label{fig:mot_attr}
\end{figure}

We further analyze the influence of different pedestrian properties, including: (a) tracking duration; (b) tracking distance; (c) moving speed; (d) scale (height); (e) scale variation (the standard deviation of height); (f) occlusion. For each property, we divided the pedestrian targets into 3 subsets from easy to hard. Besides, in order to eliminate the influence of detectors, we used the ground-truth bounding boxes as input here. Fig.~\ref{fig:mot_attr}(b)(c) show that the tracking distance and moving speed are the most influential factors to trackers' performance. In Fig.~\ref{fig:mot_attr}(a), the impact of tracking duration on tracker performance is not obvious because there are many stationary or slow moving people in the scene. 

\subsection{Group Detection}

\begin{figure*}[htbp]
    \centering
    \includegraphics[width= \linewidth]{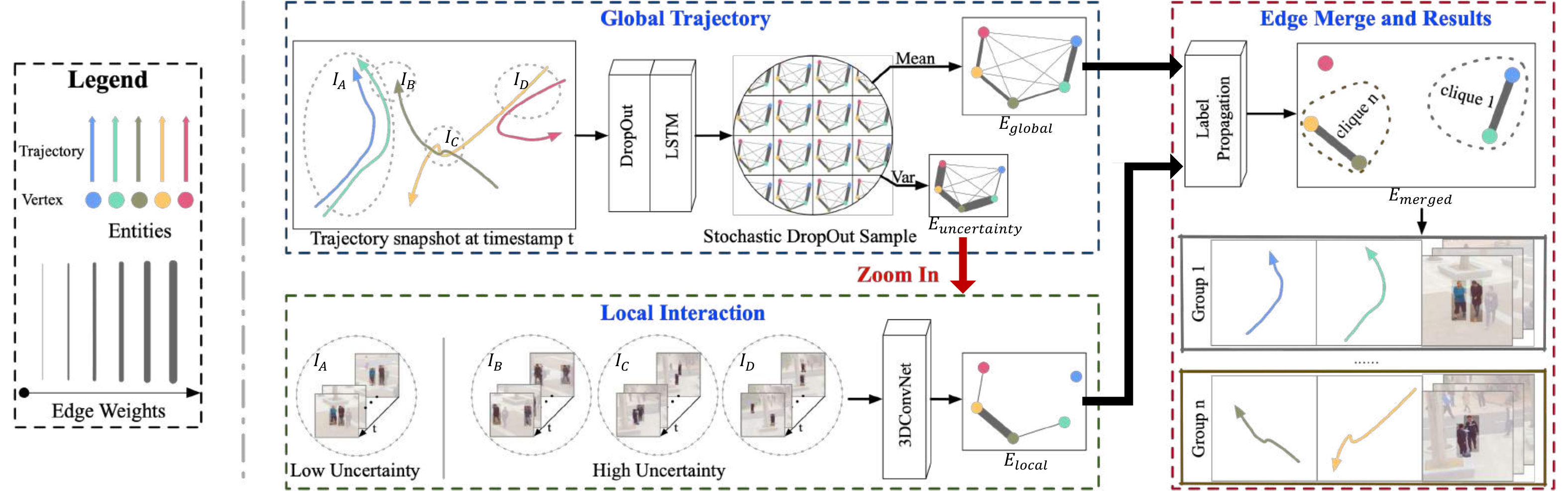}
    \caption{
     Global-to-local zoom-in framework for interaction-aware group detection. The Global Trajectory, Local Interaction, Zoom In, and Edge Merging modules are associated. Different color vertices and trajectories stand for different human entities. Line thickness represents the edge weights in the graph. \textbf{(1) Global Trajectory}: Trajectories are firstly fed into LSTM encoder with dropout layer to obtain embedding vectors and then construct a graph where the edge weight is L2 distance between embedding vectors. \textbf{(2) Zoom In}: By repeating inference with dropout activated as Stochastic Sampling~\cite{kendall2015bayesian}, $E_{global}$ and $E_{uncertainty}$ are obtained from sample mean and variance respectively. \textbf{(3) Local Interaction}: The local interaction videos corresponding to high uncertainty edges($I_{B}\sim I_{D}$)are further checked using video interaction classifier (3DConvNet~\cite{hara3dcnns}). \textbf{(4) Edge Merge and Results}: Edges are merged using label propagation~\cite{zhan2018consensus}, and cliques remaining in the graph are the group detection results.
    }
    \label{fig:group_pipeline}
\end{figure*}

Group detection aims at identifying groups of people from crowds. Unlike existing datasets that focus on either the similarity of global trajectories \cite{pellegrini2009you} or the stability of local spatial structure \cite{choi2014discovering}, the advance of PANDA with joint wide-FoV global information, high-resolution local details and temporal activities imposes rich information for group detection. 

Furthermore, as indicated by the recent advances on trajectory embedding \cite{gao2017identifying,co2018self}, trajectory prediction \cite{chandra2019traphic,choi2019learning} and interaction modeling in video recognition \cite{ibrahim2016hierarchical,shu2017cern,feichtenhofer2018slowfast}, these tasks are strongly correlated to the group detection task. For example, modeling group interaction can help improve the trajectory prediction performance \cite{xu2018encoding,chandra2019traphic,choi2019learning}, while learning a good trajectory embedding is also beneficial for video action recognition \cite{wang2015action,gao2017identifying,co2018self}. However, none of previous research has investigated how those multi-modal information can be incorporated into the group detection task. 
Hence, we propose the \emph{interaction-aware group detection task}, where video data and multi-modal annotations (spatial-temporal human trajectory, face orientation, and human interaction) are provided as input for group detection.

\textbf{Framework.} We further design a global-to-local zoom-in framework as shown in Fig.~\ref{fig:group_pipeline} to validate the incremental effectiveness of local visual clues to global trajectories.
More specifically, human entities and their relationships are represented as vertices and edges respectively in graph $G=(V,E)$. And features from multiple scales and modalities such as the global trajectory, face orientation vector, and local interaction video are used to generate edge set $E_{global}$ and $E_{local}$. Following a global-to-local strategy \cite{najibi2018autofocus,gao2018dynamic,li2019zoom,chen2019collaborative}, $E_{global}$ is firstly obtained by calculating L2 distance in feature space for each trajectory embedding vector, which comes from LSTM encoder like common practice \cite{gao2017identifying}. After that, uncertainty-based \cite{kendall2015bayesian,kendall2017uncertainties} and random selection policies are adopted to determine the sub-set of edges that need to be further checked using visual clues. Then, video interaction scores among entities are estimated by spatial-temporal ConvNet~\cite{hara3dcnns}.
The combinations of obtained edge sets, e.g., $E_{global}\cup E_{local}$ or $E_{global}$, are merged using label propagation \cite{zhan2018consensus}, and the  cliques remaining in the graph are the group detection results. Finally, we can estimate the incremental effectiveness with the performance metrics specified in \cite{choi2014discovering} under different combinations. 
\textbf{Global Trajectory.} To obtain the global trajectory edge set $E_{global}$ and edge weight function $w_{global}: E_{global} \rightarrow \mathbb{R}$, we use a simple LSTM(4 layers,128 hidden state) and embedding learning with triplet loss(margin=0.5) to extract the sequence embedding vector for each vertex $v$(denoted as $F_{v} \in \mathbb{R}^{512}, v \in V$). And then the edge weight function is calculated by:
\begin{equation}
   w_{global}(e)=||F_u-F_v||_2 ,
\end{equation}
where $e=\{u,v\}$ and $\ u,v \in V$.

More specifically about embedding network, the input trajectory is the variable-length sequence where each element $\in \mathbb{R}^6$ consists of bounding box coordinates(4 scalar), face orientation angle(1 scalar,optional), and timestamp(1 scalar). The output $F_{v}$ is obtained by concatenating the hidden state vector and cell output vector in LSTM. The supervision signal is given by triplet loss which enforces trajectories from the same group to have small L2 distance in embedding feature space and trajectories from different groups to have a large distance.

\textbf{Local Interaction.} As mentioned in the paper, calculating interaction score for each pair of human entities is inefficient and we only check a subset of entity pairs. In other words, given that $E_{local} \subset E_{global},|E_{local}|<<|E_{global}|$, $w_{local}: E_{local} \rightarrow \mathbb{R}$ is the target. More specifically, for each $e \in E_{local}$, several local video candidate clips $clip_{e}=\{clip_{e,i}\}$ is firstly cropped spatially and temporally from full video by filtering using the relative distance between 2 entities which is possible for interaction. The $clip_{e,i}$ is the variable-length sequence where each frame$\in \mathbb{R}^{4\times H\times W}$ consists of 3 channel RGB image and 1 channel interaction persons mask.And then for each  $clip_{e,i}$ we use Spatial-temporal 3D ConvNet\cite{hara3dcnns} as local video classifier which estimates the interaction score. Finally, $w_{local}$ is obtained by averaging interaction score of all the $clip_{e}$ as follow:

\begin{equation}
w_{local}(e)=\frac{\sum_{i=0}^{N_{clips}}(ConvNet(clip_{e,i}))}{N_{clips}},
\end{equation}
where $e \in E_{local}$ and $N_{clips}$ denotes the number of clips.

We use pre-trained weight from large scale dataset Kinetics\cite{kay2017kinetics}  and follows the same hyper-parameter, loss function as \cite{hara3dcnns}.

\begin{figure}[htbp]
    \centering
    \includegraphics[width=\linewidth]{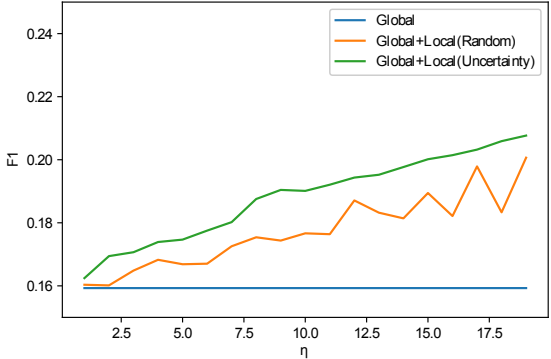}
    \caption{Sensitivity study of $\eta$(average on $\tau$). As $\eta$ increase, performances of all three model are improved and computation consumption increase as well. However, using global feature, local feature and uncertainty can achieve higher performance than random zoom in policy or without local feature under different $\eta$ value.}
    \label{fig:sensitivity_eta}
\end{figure}

\begin{figure}[htbp]
    \centering
    \includegraphics[width=\linewidth]{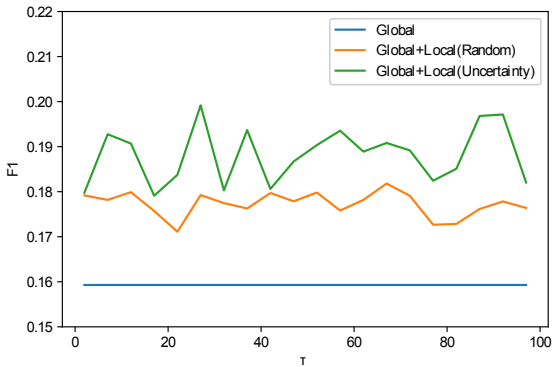}
    \caption{Sensitivity study of $\tau$(average on $\eta$). Using global feature, local feature and uncertainty can achieve higher performance than random policy or without local feature. And there is no significant increase in performance as $tau$ increasing from 10 to 510.}
    \vspace{-5mm}
    \label{fig:sensitivity_tau}
\end{figure}

\textbf{Zoom-in policy.} Zoom-in module solves the problem of selecting a subset of edge $E_{uncertainty}$ to calculate local interaction scores given $E_{global}$. And each edge $e \in E_{uncertainty}$ is further fed into the local interaction module and then $E_{local},w_{local}$ are obtained as above. There are 2 methods compared in the paper: random selection and uncertainty-based method. For the former one, $E_{uncertainty}$ consists of $\eta$ samples which are randomly selected from $E_{global}$ and predicted to be positive. For the latter one, the top $\eta$ positive predicted uncertain edges are selected. To estimate the uncertainty, stochastic dropout sampling\cite{kendall2015bayesian} is adopted. More specifically, with dropout layer activated and perform inference $\tau$ times per input. Thus for each edge score there are $\tau$ estimations and we can use the variance among the estimations as the desired uncertainty. Further more, the performance sensitivity study of $\eta$ and $\tau$ is shown in Fig.~\ref{fig:sensitivity_eta} and Fig.~\ref{fig:sensitivity_tau}

\textbf{Edge Merging Strategy.} Given $E_{global}$, $E_{local}$ and $w_{global}$, $w_{local}$ defined on them, label propagation strategy\cite{zhan2018consensus} is adopted to delete or merge edges with adaptive threshold in a iterative manner. While edges are gradually deleted, the graph is divided into several disconnected components which is the group detection result.

\textbf{Trajectory source in group detection.} We encourage users to explore the integrated solution which takes MOT result trajectory as group detection input. However, in our experiment, even the SOTA MOT method can not address the serious ID-switch, trajectory fragmentation problem. Thus, we separate the MOT task and group detection task for the first step benchmark and the previous incremental effectiveness experiment. The released dataset provides sufficient annotation and we encourage users to explore the more robust MOT methods or the integrated solution of 2 tasks. As a result of using trajectory annotation, the training-testing set split is different from previous task. In the group detection task, we use Training set in Tab.~\ref{tab:sta_video} to train and test. More specifically, scene \emph{University Canteen} is used as the testing set and the rest 8 scenes are used as training sets.

\textbf{Results.} Experimental results are shown in Tab.~\ref{tab:group_effectiveness}. Half metrics \cite{choi2014discovering} including precision, recall, and F1 where group member $IoU=0.5$ are used for evaluation.
The performance is improved significantly by leveraging $E_{local}$ as well as uncertainty estimation, which further validates the effectiveness of local visual clues provided by PANDA.

%


\begin{table}
\begin{center}
\resizebox{\columnwidth}{!}{%
\begin{tabular}{|c|c|c|c|c|}
\hline
Edge Sets & Zoom In & Precision & Recall & F1 \\ 
\hline\hline
$E_{global}$ & / & 0.237 & 0.120 & 0.160 \\
$E_{global} \cup E_{local}$ & Random & 0.244 & 0.133 & 0.172 \\
$E_{global} \cup E_{local}$ & Uncertainty & 0.293 & 0.160 & 0.207 \\ \hline
\end{tabular}
}
\end{center}
\caption{
Incremental Effectiveness (half metric \cite{choi2014discovering}). The random zoom-in policy randomly selects several local videos to estimate interaction score while the uncertainty-based one selects local videos depending on the uncertainty estimation from Stochastic Dropout Sample \cite{kendall2015bayesian}.}
\vspace{-5mm}
\label{tab:group_effectiveness}
\end{table}

\section{Conclusion}


In this paper, we introduced a gigapixel-level video dataset (PANDA) for large-scale, long-term, and multi-object human-centric visual analysis. The videos in PANDA are equipped with both wide FoV and high spatial resolution. Rich and hierarchical annotations are provided. We benchmarked several state-of-the-art algorithms for the fundamental human-centric tasks, pedestrian detection and tracking. The results demonstrate that they are heavily challenged for accuracy due to the significant variance of pedestrian pose, scale, occlusion and trajectory, etc., and efficiency due to the large image size and the huge amount of objects in single frame. Besides, we introduced a new task, termed as interaction-aware group detection based on the characteristics of PANDA. We proposed a global-to-local zoom-in framework which combines both global trajectories and local interactions, yielding promising group detection performance. Based on PANDA, we believe the community will develop new effective and efficient algorithms for understanding complicated behaviors and interactions of crowd in large-scale real-world scenes.





\newcommand{\tabincell}[2]{\begin{tabular}{@{}#1@{}}#2\end{tabular}}

\section{Appendix}

\subsection{Statistical Overview of Scenes and Label Description}

\begin{figure*}[htbp]
\centering
\includegraphics[width=\linewidth]{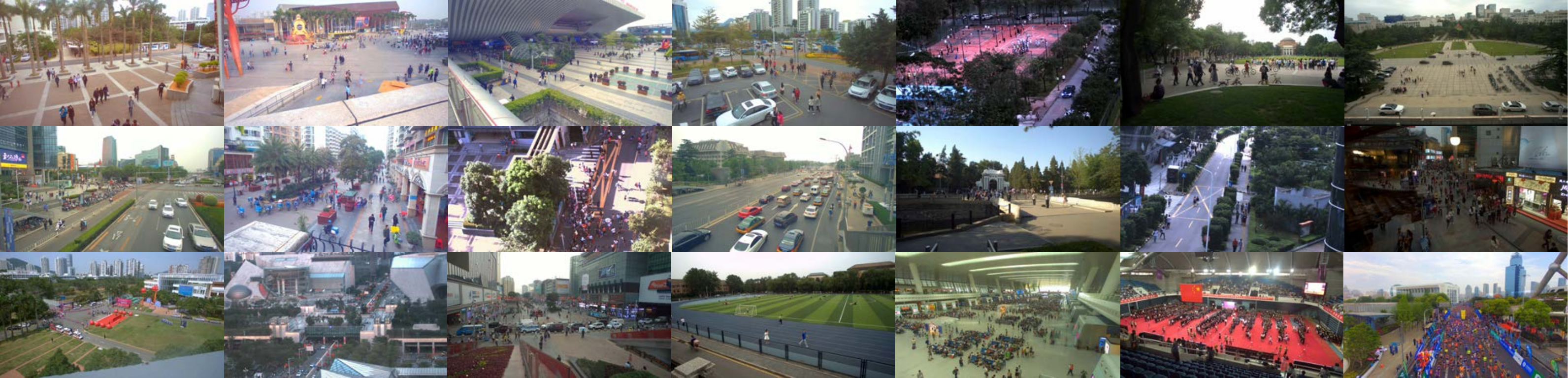}
\caption{Overview of 21 real-world outdoor scenes in PANDA.}
\vspace{-3mm}
\label{fig:scenes}
\end{figure*}

Currently, PANDA consists of 21 real-world large-scale scenes, as shown in Fig.~\ref{fig:scenes}, and the annotation details are illustrated in Tab.~\ref{tab:anno}. We are continuously collecting more videos to enrich our dataset. Note that all the data was collected in public areas where photography is officially approved, and it will be published under the Creative Commons Attribution-NonCommercial-ShareAlike 4.0 License \cite{license4.0}.

In Tab.~\ref{tab:sta_img} and Tab.~\ref{tab:sta_panda_c}, we give an overview of the training and testing set characteristics for PANDA and PANDA-Crowd images, respectively. In Tab.~\ref{tab:sta_video}, we give an overview of the training and testing set characteristics for PANDA videos.

\begin{table*}[!htbp]
\footnotesize
\resizebox{\textwidth}{!}{
\begin{tabular}{|c|c|c|c|}
\hline
Data & \multicolumn{2}{c|}{Attributes} & Labels \\ \hline \hline
\multirow{7}{*}{Image}  & \multirow{3}{*}{Location} & Person ID & --\\  
 & & Head Point & Marked in the geometric center of the human head \\ 
 & & Bounding Box & Estimated Full Body; Visible Body; Head \\ \cline{2-4}  
 & \multirow{4}{*}{Properties} & Age & Child; Adult \\ 
 & & Posture & Walking; Standing; Sitting; Riding; Held in Arms \\ 
 & & Rider Type & Bicycle Rider; Tricycle Rider; Motorcycle Rider \\  
 & & Special Cases & Fake Person; Dense Crowd; Ignore \\ \hline
\multirow{12}{*}{Video} & \multirow{2}{*}{Trajectories} & Person ID & -- \\
 & & Bounding Box & Visible Body; Estimated Full Body (for disappearing case) \\ \cline{2-4}
 & \multirow{4}{*}{Properties} & Age & Child; Youth and Middle-aged; Elderly \\ 
 & & Gender & Male; Female \\ 
 & & Face Orientation & $\uparrow \downarrow  \rightarrow  \leftarrow \nearrow  \searrow  \swarrow  \nwarrow$\\ 
 & & Occlusion Degree & W/O Occlusion; Partial Occlusion; Heavy Occlusion; Disappearing \\ \cline{2-4}
 & \multirow{3}{*}{Group} & Group ID & -- \\ 
 & & Intimacy & Low; Middle; High \\ 
 & & Group Type & Acquaintance; Family; Business \\ \cline{2-4}
 & \multirow{3}{*}{Interaction} & Begin/End Frame  & -- \\
 & & Interaction Type & Physical Contact; Body Language; Face Expressions; Eye Contact; Talking\\ 
 & & Confidence Score & Low; Middle; High \\ \hline
\end{tabular}
}
\caption{Annotation Details in PANDA dataset.}
\label{tab:anno}
\end{table*}

\begin{table*}[!htbp]
\resizebox{\textwidth}{!}{
\begin{tabular}{|c|c|c|c|c|c|c|c|c|}
\hline
Scene & \#Sub-scene & \#Image & Resolution & \tabincell{c}{Mean\\\#Person} & \tabincell{c}{Mean\\\#Special Case} & \tabincell{c}{Mean\\Person Height} & \tabincell{c}{Mean\\Occlusion Ratio} & Camera Height \\ \hline \hline
\multicolumn{9}{|c|}{Training Set} \\ \hline
University Canteen & 1 & 30 & 26753$\times$15052 & 52.7 & 23.7 & 906.79 & 0.11 & 2nd Floor \\
Xili Crossroad & 1 & 30 & 26753$\times$15052 & 174.2 & 27.3 & 506.78 & 0.13 & 2nd Floor \\ 
Train Station Square & 2 & 15/15 & 26583$\times$14957 & 272.1 & 75.8 & 328.05 & 0.11 & 2nd Floor \\ 
Grant Hall & 1 & 30 & 25306$\times$14238 & 133.1 & 22.6 & 583.38 & 0.13 & 1st Floor \\ 
University Gate & 1 & 30 & 26583$\times$14957 & 122.6 & 43.9 & 617.88 & 0.20 & 1st Floor \\ 
University Campus & 1 & 30 & 26088$\times$14678 & 223.0 & 26.7 & 293.80 & 0.08 & 8th Floor \\ 
East Gate & 1 & 30 & 25831$\times$14533 & 175.7 & 37.4 & 201.43 & 0.14 & 2nd Floor \\ 
Dongmen Street & 1 & 30 & 25151$\times$14151 & 289.4 & 79.4 & 551.16 & 0.15 & 2nd Floor \\
Electronic Market & 1 & 30 & 25306$\times$14238 & 571.6 & 113.4 & 339.17 & 0.23 & 2nd Floor \\
Ceremony & 1 & 30 & 25831$\times$14533 & 250.3 & 51.3 & 308.69 & 0.11 & 5th Floor \\ 
Shenzhen Library & 2 & 15/15 & \tabincell{c}{32129$\times$24096 /\\ 31746$\times$23810} & 190.9 & 59.1 & 321.77 & 0.13 & 20th Floor \\
Basketball Court & 2 & 15/15 & \tabincell{c}{31753$\times$23810 /\\ 31746$\times$23810} & 86.7 & 10.4 & 928.29 & 0.07 & 10th Floor \\ 
University Playground & 2 & 15/15 & \tabincell{c}{27098$\times$15246 /\\ 25654$\times$14434} & 127.5 & 14.4 & 307.45 & 0.04 & 2nd Floor \\ \hline
\multicolumn{9}{|c|}{Testing Set} \\ \hline
OCT Habour & 1 & 30 & 26753$\times$15052 & 278.8 & 48.5 & 495.34 & 0.10 & 2nd Floor \\ 
Nanshani Park & 1 & 30 & 32609$\times$24457 & 83.6 & 24.9 & 1,108.77 & 0.14 & 5th Floor \\ 
Primary School & 2 & 15/15 & 31760$\times$23810 & 233.9 & 24.0 & 1,096.56 & 0.08 & 19th Floor \\ 
New Zhongguan & 1 & 30 & 26583$\times$14957 & 352.6 & 85.2 & 353.08 & 0.16 & 2nd Floor \\ 
Xili Street & 2 & 30/15 & \tabincell{c}{26583$\times$14957 /\\ 26753$\times$15052} & 118.4 & 47.2 & 642.51 & 0.13 & 2nd Floor \\ \hline
\end{tabular}
}
\caption{Statistics and train-test set split for 18 scenes of PANDA images. '\#' represents 'The number of'; Sub-scene represents data captured in the same scene, but with different viewpoints or recording time; ‘Mean’ represents the mean of the value for each image; Person height is calculated in pixels; Occlusion Ratio is the ratio of the visible body bbox area to the estimated full body bbox area.}
\vspace{-1mm}
\label{tab:sta_img}
\end{table*}


\begin{table*}[!htbp]
\small
\centering
\begin{tabular}{|c|c|c|c|c|}
\hline
Scene & \#Image & Resolution & Mean \#Person & Camera Height \\ \hline
\hline
\multicolumn{5}{|c|}{Training Set} \\ \hline
Marathon & 15 & 26908$\times$15024 & 3,619.2 & 4th Floor \\
Graduation Ceremony & 15 & 26583$\times$14957 & 1,483.0 & 2nd Floor \\ \hline
\multicolumn{5}{|c|}{Testing Set} \\ \hline
Waiting Hall & 15 & 26558$\times$14828 & 3,039.1 & 2nd Floor \\ \hline
\end{tabular}
\caption{Statistics and train-test set split for 3 scenes of PANDA-Crowd images. '\#' represents 'The number of'; ‘Mean’ represents the mean of the value for each image.}
\label{tab:sta_panda_c}
\end{table*}

\begin{table*}[!htbp]
\resizebox{\textwidth}{!}{
\begin{tabular}{|c|c|c|c|c|c|c|c|c|}
\hline
Scene & \#Frame & FPS & Resolution & \#Tracks & \#Boxes & \#Groups & \#Single Person & Camera Height \\ \hline \hline
\multicolumn{9}{|c|}{Training Set} \\ \hline
University Canteen & 3,500 & 30 & 26753$\times$15052 & 295 & 335.2k & 75 & 123 & 2nd Floor \\ 
OCT Habour & 3,500 & 30 & 26753$\times$15052 & 736 & 1,270.1k & 205 & 191 & 2nd Floor \\ 
Xili Crossroad & 3,500 & 30 & 26753$\times$15052 & 763 & 1,065.0k & 163 & 393 & 2nd Floor \\ 
Primary School & 889 & 12 & 34682$\times$26012 & 718 & 465.6k & 117 & 119 & 19th Floor \\ 
Basketball Court & 798 & 12 & 31746$\times$23810 & 208 & 118.4k & 34 & 54 & 10th Floor \\ 
Xinzhongguan & 3,331 & 30 & 26583$\times$14957 & 1,266 & 1,626.0k & 186 & 857 & 2nd Floor \\ 
University Campus & 2,686 & 30 & 25479$\times$14335 & 420 & 658.6k & 83 & 123 & 8th Floor \\ 
Xili Street 1 & 3,500 & 30 & 26583$\times$14957 & 662 & 950.0k & 144 & 325 & 2nd Floor \\ 
Xili Street 2 & 3,500 & 30 & 26583$\times$14957 & 290 & 425.7k & 59 & 152 & 2nd Floor \\
Huaqiangbei & 3,500 & 30 & 25306$\times$14238 & 2,412 & 3,054.5k & 310 & 1,730 & 2nd Floor \\ \hline
\multicolumn{9}{|c|}{Testing Set} \\ \hline 
Train Station Square & 3,500 & 30 & 26583$\times$14957 & 1,609 & 1,682.7k & 178 & 1,213 & 2nd Floor \\ 
Nanshan i Park & 889 & 12 & 32609$\times$24457 & 402 & 132.6k & 78 & 199 & 5th Floor \\ 
University Playground & 3,560 & 30 & 25654$\times$14434 & 309 & 574.3k & 60 & 165 & 2nd Floor \\ 
Ceremony & 3,500 & 30 & 25831$\times$14533 & 677 & 1,444.7k & 143 & 317 & 5th Floor \\
Dongmen Street & 3,500 & 30 & 26583$\times$14957 & 1,922 & 1,676.4k & 331 & 1,170 & 2nd Floor \\ \hline
\end{tabular}
}
\caption{Statistics and train-test set split for 15 scenes of PANDA videos. '\#' represents 'The number of'; FPS represents 'Frames Per Second'.}
\label{tab:sta_video}
\end{table*}

\subsection{Evaluation Metrics}
\subsubsection{Evaluation Metrics for Object Detection}
Our evaluation metrics are the Average Precision $AP_{.50}$ and Average Recall $AR$, which are adopted from the MS COCO \cite{lin2014microsoft} benchmark. Specifically, $AP_{.50}$ is defined as the average precision at ${\rm IoU}=0.50$ and $AR$ is defined as average recall with IoU ranging in $[0.5, 0.95]$ with a stride of 0.05. To get rid of the bias towards the overcrowded frames, the maximum number of detection results on each frame is set to 500 for the calculation of AP and AR. Precision and recall is defined as follows:
\begin{equation}
    precision =\frac{{\rm TP}}{{\rm TP}+{\rm FP}}
\end{equation}
\begin{equation}
recall = \frac{{\rm TP}}{{\rm TP} + {\rm FN}}
\end{equation}
where TP, FP, FN are the number of True Positive, False Positive, False Negative, respectively.
The Interaction-of-Union (IoU) between two bounding boxes is defined as follows:
\begin{equation}
{\rm IoU}=\frac{|A \cap B|}{|A \cup B|}
\end{equation}
where $A,B$ are pixel areas of the predicted and ground-truth bounding boxes respectively.

\subsubsection{Evaluation Metrics for Multiple Object Tracking}

This section includes additional details regarding the definitions of the evaluation metrics for multiple objects tracking, which are partially explained in Section 4.2. The measurements are adopted from the MOT Challenge \cite{milan2016mot16} benchmarks. 
In MOT Challenge, 2 sets of measures are employed: The CLEAR metrics proposed by \cite{stiefelhagen2006clear}, and a set of track quality measures introduced by \cite{wu2006tracking}.

The distance measure, \emph{i.e.}, how close a tracker hypothesis is to the actual target, is determined by the intersection over union (IoU) between estimated bounding boxes and the ground truths. The similarity threshold $t_d$ for true positives is empirically set to 50\%.

The Multiple Object Tracking Accuracy (MOTA) combines three sources of errors to evaluate a tracker’s performance, defined as

\begin{equation}
   {\rm MOTA} = 1 - \frac{\sum_{t}({\rm FN}_t + {\rm FP}_t + {\rm IDSW}_t)}{\sum_{t}{\rm GT}_t}
\end{equation}
where $t$ is the frame index. FN, FP, IDSW and GT respectively denote the numbers of false negatives, false positives, identity switches and ground truths. The range of MOTA is ($-\infty$, 1], which becomes negative when the number of errors exceeds the ground truth objects.

Multiple Object Tracking Precision (MOTP) is used to measure misalignment between annotated and predicted object locations, defined as

\begin{equation}
   {\rm MOTP} = 1 - \frac{\sum_{t,i}d_{t,i}}{\sum_{t}c_t}
\end{equation}
where $c_t$ denotes the number of matches in frame $t$ and $d_{t,i}$ is the bounding box overlap of target $i$ with its assigned ground truth object. MOTP thereby gives the average overlap between all correctly matched hypotheses and their respective objects and ranges between $t_d$ := 50\% and 100\%. 
According to \cite{milan2016mot16}, in practice, it mostly quantifies the localization accuracy of the detector, and therefore, it provides little information about the actual performance of the tracker.

\subsubsection{Evaluation Metrics for Group Detection}
As discussed in \cite{choi2014discovering}, the half metric refers to a single detected group prediction that is positive if the detected group contains at least half of the elements of the Ground Truth group (and vice-versa). And then we can calculate precision, recall, and F1 based on the positive and negative samples. More specifically, each detected group ($Grp_{pd}$) as well as ground truth($Grp_{gt}$) is a set of group member:
\begin{equation}
    Grp_{*}=\{v| v \in V\ and\ v\ belong\ to\ the\ group\}
\end{equation}
And one detected group is regarded as correct under half metric if and only if it satisfy the following:
\begin{equation}
    \frac{Grp_{pd}\cap Grp_{gt}}{max(|Grp_{pd}|,|Grp_{gt}|)}
    >0.5
\end{equation}

\end{document}